\documentclass[10pt, a4paper]{article}

\usepackage[]{lrec-coling2024} 

\usepackage{booktabs}
\usepackage{amssymb}
\usepackage{bm}
\usepackage{amsfonts}
\usepackage{amsmath}
\usepackage{cases}
\usepackage{multirow}
\usepackage{mathtools}
\usepackage{color}
\usepackage{stmaryrd}
\usepackage{makecell}
\usepackage{graphicx,subfig}
\usepackage{pifont}
\usepackage{changes}
\usepackage{marvosym}

\title{TransERR: Translation-based Knowledge Graph Embedding via Efficient Relation Rotation}

\name{Jiang Li$^{1,2,3}$, Xiangdong Su$^{1,2,3}$\textsuperscript{(\Letter)}, Fujun Zhang$^{1,2,3}$, Guanglai Gao$^{1,2,3}$}

\address{$^1$College of Computer Science, Inner Mongolia University, Hohhot, China\\ 
$^2$National \& Local Joint Engineering Research Center of \\Intelligent Information Processing Technology for Mongolian \\
$^3$Inner Mongolia Key Laboratory of Mongolian Information Processing Technology \\
lijiangimu@gmail.com, cssxd@imu.edu.cn,32209093@mail.imu.edu.cn, csggl@imu.edu.cn\\}

\abstract{
This paper presents a translation-based knowledge geraph embedding method via efficient relation rotation (TransERR), a straightforward yet effective alternative to traditional translation-based knowledge graph embedding models. Different from the previous translation-based models, TransERR encodes knowledge graphs in the hypercomplex-valued space, thus enabling it to possess a higher degree of translation freedom in mining latent information between the head and tail entities. To further minimize the translation distance, TransERR adaptively rotates the head entity and the tail entity with their corresponding unit quaternions, which are learnable in model training. We also provide mathematical proofs to demonstrate the ability of TransERR in modeling various relation patterns, including symmetry, antisymmetry, inversion, composition, and subrelation patterns. The experiments on 10 benchmark datasets validate the effectiveness and the generalization of TransERR. The results also indicate that TransERR can better encode large-scale datasets with fewer parameters than the previous translation-based models. Our code and datasets are available at~\url{https://github.com/dellixx/TransERR}.
 \\ \newline  }

\begin{document}

\maketitleabstract

\section{Introduction}
Knowledge graphs (KGs), also known as semantic networks, represent networks of real-world entities (objects, events, concepts, etc.) and describe the associations between them. In fact, KGs contain factual triplets $(head, relation, tail)$, which are denoted as $(h, r, t)$. Several open source KGs are available, including FreeBase~\cite{4bollacker2008freebase}, DBpedia~\cite{45lehmann2015dbpedia} and  NELL~\cite{46mitchell2018never}. They facilitate the development of downstream tasks, such as question answering~\cite{2}, semantic search~\cite{3junior2020knowledge} and relation extraction~\cite{1}. However, there is a problem with missing links in KGs. Therefore,  knowledge graph embedding (KGE) task has recently received considerable attention.

The mainstream approaches are to learn low-dimensional representations of entities and relations and utilize them to predict new facts. Most of them learn the embeddings of KGs based on score functions of the translational distance between the head and tail entities. Weak degree of freedom (\ding{56}) and Translational Transformations (\ding{52}). RotatE~\cite{31sun2019rotate}, PairRE~\cite{52chao2020pairre} and TranSHER~\cite{li-etal-2022-transher} have demonstrated that translational transformations better capture the relations between entities, thereby enhancing the expressive power of KG embeddings. However, these models are constrained by the degrees of freedom in the transformations, implying that they are not fully expressive.

Recently, QuatE~\cite{zhang2019quaternion} and Rotat3D~\cite{15gao2020rotate3d} have demonstrated that vectors exhibit a higher degree of rotational freedom in the quaternion space.  This implies that one can attain a greater level of rotational freedom in representing when compared to traditional vector spaces. Nevertheless, QuatE and Rotat3D exclusively employ rotational transformations within quaternion space, thus overlooking translational changes. This limitation leads to a reduced capacity for modeling spatial transformations, resulting in a decrease in performance of link prediction. High degree of freedom (\ding{52}) and Translational Transformations (\ding{56}).

Based on the above facts, this paper proposes a KGE method named TransERR that combines a high degree of rotational freedom (\ding{52}) and translational transformations (\ding{52}). TransERR encodes knowledge graphs in the hypercomplex-valued space and utilizes two unit quaternion vectors to rotate the head entity and the tail entity, respectively. The unit quaternion vectors are learnable in model training, and they are highly to smooth rotation and spatial translation in the hypercomplex-valued vector space. In addition, two unit quaternion rotation vectors can further narrow the translation distance between the head and tail entities. As a result, TransERR possesses a higher degree of translation freedom in mining latent information between the head and tail entities. We evaluate the effectiveness and generalisation of our model on 10 KG benchmark datasets of different sizes. The experimental results show that TransERR significantly outperforms the existing state-of-the-art distance-based models.  We provide mathematical proofs to demonstrate that TransERR can infer key relation patterns simultaneously. Furthermore, we demonstrate that TransERR can better model complex relations of KGs than the existing distance-based models and surpass the baselines on large-scale datasets with fewer parameters.

\section{Related Work}
Most existing KGE models can be roughly classified into two categories: Translation Models and Semantic Matching Models~\cite{wang2017knowledge,ji2021survey}. The former measures the plausibility of a fact as a translation distance between two entities, while the latter measures the plausibility of facts by matching latent semantics of entities and relations. In this paper, TransERR belongs to translation models.

\textbf{Translation Models.} Translation-based models, also known as distance-based models. Motivated by the translation invariance in word2vec \cite{19mikolov2013efficient}, TransE \cite{7bordes2013translating}  defines the distance between $h + r$ and $t$  with the $l_1$ or $l_2$ norm constraint. After that, TransH \cite{81wang2014knowledge}, TransR \cite{9lin2015learning} and TransD \cite{10ji2015knowledge}  employ different projection strategies to adjust graph embeddings. TranSparse \cite{20ji2016knowledge} overcomes heterogeneity and imbalance by combining TranSparse (share) and TranSparse (separate). TransMS \cite{21yang2019transms} perform multi-directional semantic transfer with non-linear functions and linear deviation vectors. RotatE \cite{31sun2019rotate} defines each relation as a rotation for a triplet from the head entity to the tail entity. Rotat3D \cite{15gao2020rotate3d} maps entities into 3D space and defines relations similar to RotatE. However, Rotat3D is limited by optimizing the translation distance with only relation embedding information. Recently, PairRE~\cite{52chao2020pairre}, TripleRE~\cite{yu2022triplere} and TranSHER~\cite{li-etal-2022-transher} employ multiple relations to improve the degree of freedom for relational rotation. Different from previous work, TransERR processes a high degree of rotation in the quaternion space.  Benefiting from our model structure, TransERR can model all important relation patterns simultaneously and allow for better optimization of the translation distance between entities and, thus better mine the latent information.

\textbf{Semantic Matching Models.}
Semantic matching  models mine possible semantic associations between entities and relations. The RESCAL~\cite{27nickel2011three} represents each relation as a non-singular matrix. DistMult~\cite{29yang2014embedding} uses a diagonal matrix rather than a non-singular matrix to address the problem of the excessive number of parameters in RESCAL. ComplEx~\cite{14trouillon2016complex} introduces complex-valued spaces into the KGs. TuckER~\cite{TuckER2019tucker} employs Tucker decomposition of a binary tensor to model a KG. SimplE$^+$~\cite{58fatemi2019improved} extends SimplE~\cite{kazemi2018simple} and focuses on encoding the subrelation pattern. QuatE~\cite{zhang2019quaternion}, QuatRE~\cite{nguyen2022quatre} and QuatSE~\cite{li2022quatse} take advantage of quaternion representations to enable rich interactions between entities and relations. Deep neural networks have received a great deal of attention in recent years, e.g., ConvE~\cite{dettmers2018convolutional}, ConvKB~\cite{nguyen2017novel}, InteractE~\cite{vashishth2020interacte} and R-GCNs~\cite{schlichtkrull2018modeling}. However, they are difficult to analyze as they work as a black box.

\begin{figure*}[!htb]
	\centering
	\includegraphics[width=1.9\columnwidth]{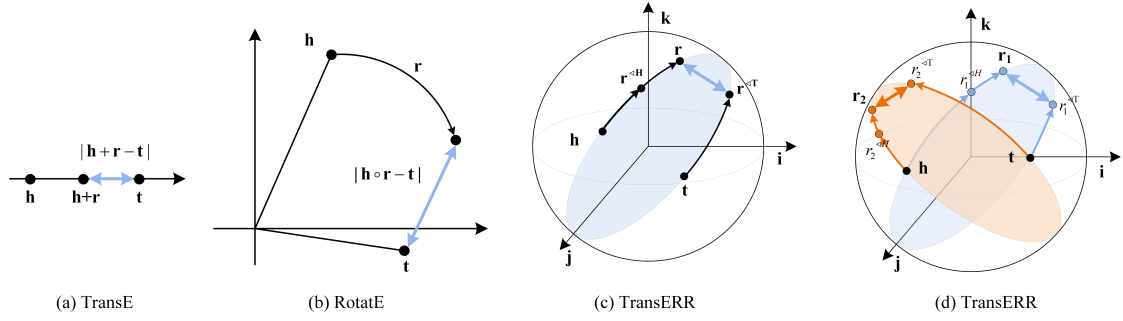}
	\caption{Illustration of TransE, RotatE and TransERR. TransE, RotatE and TransERR encode knowledge graphs in the real-valued space, complex-valued space and hypercomplex-valued space, respectively. $\circ$ denotes Hadamard product.  The distance function of TransERR is $\parallel \mathbf{h} \otimes \mathbf{r^{\vartriangleleft H}} + \mathbf{r}- \mathbf{t} \otimes \mathbf{r^{\vartriangleleft T}}   \parallel$.}
	\label{fig:transqr}
\end{figure*}

\section{Background and Notation}\label{sec3}

In this section, we first introduce the link prediction task. Secondly, we introduce the key relation patterns and complex relation patterns that are commonly studied in this task. Finally, we provide a brief introduction to quaternion algebra.

\textbf{Link Prediction.}
A knowledge graph is usually described as $\mathcal{G}=( \mathcal{E},\mathcal{R},\mathcal{T} )$, where $\mathcal{E} $,  $\mathcal{R} $ and $\mathcal{T}$ denote the set of entities, relations and triplets $(h,r,t)$, respectively. Specifically, given $(h,r,?)$, link prediction is to predict the tail entity in the triplet. Alternatively, given $(?,r,t)$, the task is to predict the head entity. The existing predicting models calculate the score function $f_r(h,t)$ and expect that the scores of correct triplets are higher than those of invalid triplets.

\textbf{Key Relation Patterns.}
In this part, we introduce several important relation patterns that have been extensively studied in link prediction task. Following \citet{58fatemi2019improved} and \citet{31sun2019rotate}, relations can also be summarized by several patterns:
\begin{itemize}
	\item \textbf{Symmetry} If $ (e_1,r,e_2)\in\mathcal{T},\forall e_1, e_2 \in \mathcal{E} \Rightarrow  (e_2,r,e_1)\in\mathcal{T}$.
	\item \textbf{Antisymmetry} If $ (e_1,r,e_2)\in\mathcal{T},\forall e_1, e_2 \in \mathcal{E} \Rightarrow (e_2,r,e_1)\notin \mathcal{T} $.
	\item \textbf{Inversion} If $ (e_1,r_1,e_2)\in\mathcal{T},\forall e_1, e_2 \in \mathcal{E} \Rightarrow (e_2,r_2,e_1)\in \mathcal{T}$.
	\item \textbf{Composition} If $ (e_1,r_1,e_2)\in\mathcal{T} \land (e_2,r_2,e_3)\in\mathcal{T},   \forall e_1, e_2, e_3 \in \mathcal{E} \Rightarrow (e_1,r_3,e_3)\in \mathcal{T}$.
	\item \textbf{Subrelation} If $ (e_1,r_1,e_2)\in\mathcal{T}, \forall e_1, e_2 \in \mathcal{E} \Rightarrow (e_1,r_2,e_2)\in\mathcal{T}$. $ (e_1,r_1,e_2)\rightarrow (e_1,r_2,e_2)$ and $r_2$ can be considered as a subrelation of $r_1$.
\end{itemize}

\textbf{Complex Relation Patterns.} We take the definition of complex relations from \citet{81wang2014knowledge}, and we calculate average number of tails per head ($tph_r$) and average number of heads per tail ($hpt_r$). If $tph_r<1.5 $ and $hpt_r <1.5 $, r is defined as one-to-one (\textbf{1-1}). If $tph_r > 1.5$ and $hpt_r> 1.5$, r is defined as many-to-many (\textbf{N-N}). If $tph_r>1.5 $ and $hpt_r < 1.5 $, r is defined as one-to-many (\textbf{1-N}). If $tph_r<1.5 $ and $hpt_r > 1.5 $, r is defined as many-to-one (\textbf{N-1}).

\textbf{Quaternion Algebra.} The quaternion is an extension of the complex number in the four-dimensional space. It consists of a real part and three imaginary parts, which is proposed by William Rowan Hamilton \cite{34hamilton1844theory}.
A quaternion $q$ is defined as $q=a+bi+cj+dk$, where $a$ is real unit and $i$, $j$, $k$ are three imaginary units. In addition, $i^2=j^2=k^2=ijk=-1$.

\begin{itemize}
	\item The unit quaternion $q^\vartriangleleft $ is defined as
	      \begin{align}
		      q^\vartriangleleft=\frac{a+bi+cj+dk}{\sqrt{a^2+b^2+c^2+d^2}}.
	      \end{align}

	\item Hamilton product of two quaternions is
	      \begin{equation}\label{eq2}
		      \begin{split}
			      q_1\otimes q_2 = &(a_1a_2-b_1b_2-c_1c_2-d_1d_2)+\\
			      &(a_1b_2+b_1a_2+c_1d_2-d_1c_2)i+\\
			      &(a_1c_2-b_1d_2+c_1a_2+d_1b_2)j+\\
			      &(a_1d_2+b_1c_2-c_1b_2+d_1a_2)k.
		      \end{split}
	      \end{equation}
\end{itemize}

\section{Methodology}

\subsection{TransERR}\label{sec4.1}
Quaternions enable expressive rotation in the hypercomplex-valued space and have more degree of freedom than translation in the real-valued space. Hence, to obtain a greater degree of translation freedom in the link prediction, TransERR encodes knowledge graphs in the hypercomplex-valued space. In addition, we rotate head entity $\mathbf{h}$ and tail entity $\mathbf{t}$ via two unit quaternion vectors $\mathbf{r^{\vartriangleleft H}}$ and $\mathbf{r^{\vartriangleleft T}}$, as shown in Figure~\ref{fig:transqr}. Unlike TransE and RotatE, TransERR utilizes Hamilton product $\otimes$ rather than Hadamard product $\circ$ in project operation to better capture  the underlying semantic features between the head entity and the tail entity embeddings. Firstly, given a triplet $(h,r,t)$, we encode $h$, $r$ and $t$ in the quaternion space, which are denoted as
\begin{equation}
	\begin{aligned}
		\mathbf{h} & = \mathbf{a_h}+\mathbf{b_h}i+\mathbf{c_h}j+\mathbf{d_h}k   \\
		\mathbf{r} & = \mathbf{a_r}+\mathbf{b_r}i+\mathbf{c_r}j+\mathbf{d_r}k   \\
		\mathbf{t} & = \mathbf{a_t}+\mathbf{b_t}i+\mathbf{c_t}j+\mathbf{d_t}k,
	\end{aligned}
\end{equation}
where $\mathbf{h},\mathbf{r},\mathbf{t} \in\mathbb{H}^{d}$, $\mathbf{a_h},\mathbf{b_h}, \mathbf{c_h}, \mathbf{d_h} \in \mathbb{R}^{\frac{d}{4}}$, $\mathbf{a_r},\mathbf{b_r}, \mathbf{c_r}, \mathbf{d_r} \in \mathbb{R}^{\frac{d}{4}}$ and $\mathbf{a_t},\mathbf{b_t}, \mathbf{c_t}, \mathbf{d_t} \in \mathbb{R}^{\frac{d}{4}}$. Next, we define two additional quaternion vectors to  rotate the head and tail entity, which are represented as $\mathbf{r^H}$ and $\mathbf{r^T}$, where $\mathbf{r^H}$ and $\mathbf{r^T} \in\mathbb{H}^{d}$. Then, we normalize these two additional quaternions ($\mathbf{r^H}$ and $\mathbf{r^T}$) to the unit quaternions ($\mathbf{r^{\vartriangleleft H}}$ and $\mathbf{r^{\vartriangleleft T}}$) to eliminate the scaling effect. They are denoted as

\begin{equation}
	\begin{aligned}
		\mathbf{r^{\vartriangleleft H}}=\frac{\mathbf{a_{r^H}}+\mathbf{b_{r^H}}i+\mathbf{c_{r^H}}j+\mathbf{d_{r^H}}k}{\sqrt{\mathbf{a_{r^H}^2}+\mathbf{b_{r^H}^2}+\mathbf{c_{r^H}^2}+\mathbf{d_{r^H}^2}}} \\
		\mathbf{r^{\vartriangleleft T}}=\frac{\mathbf{a_{r^T}}+\mathbf{b_{r^T}}i+\mathbf{c_{r^T}}j+\mathbf{d_{r^T}}k}{\sqrt{\mathbf{a_{r^T}^2}+\mathbf{b_{r^T}^2}+\mathbf{c_{r^T}^2}+\mathbf{d_{r^T}^2}}}.
	\end{aligned}
\end{equation}

The normalized operation ensures the stability of entity rotation in the quaternion space. The unit quaternion vectors are highly desirable to smooth rotation and spatial translation in the hypercomplex-valued space. In addition, two unit quaternion rotation vectors can further narrow the translation distance between the head and tail entities. Finally, we employ $\mathbf{r^{\vartriangleleft H}}$ and $\mathbf{r^{\vartriangleleft T}}$ to rotate head entity $\mathbf{h}$ and tail entity $\mathbf{t}$, respectively. Specifically, we use Hamilton product to achieve rotation operation since Hamilton product makes quaternion more expressive at rotational capability. For each triplet $(h, r, t)$, we define the distance function of TransERR as

\begin{equation}
	d_r(\mathbf{h},\mathbf{t}) = \parallel \mathbf{h} \otimes \mathbf{r^{\vartriangleleft H}} + \mathbf{r}- \mathbf{t} \otimes \mathbf{r^{\vartriangleleft T}}   \parallel.
\end{equation}

The score function $f_r(\mathbf{h},\mathbf{t}) = - d_r(\mathbf{h},\mathbf{t})$ and $\otimes$ is defined in Section~\ref{sec3}. Following \citet{31sun2019rotate}, we employ the self-adversarial negative sampling loss for TransERR, which is defined as

\begin{equation}
	\begin{aligned}
		\mathcal{L} = -\log \sigma (\gamma -d_r(\mathbf{h},\mathbf{t})) \\ - \sum_{i = 1}^{n}p(h'_i,r,t'_i)\log\sigma(d_r(\mathbf{h'_i},\mathbf{t'_i})-\gamma),
	\end{aligned}
\end{equation}
where $\sigma$ is the sigmoid function, and $\gamma$ is a fixed margin. $(h'_i,r,t'_i)$ and $d_r(\mathbf{h'_i},\mathbf{t'_i})$ represent the $i$-th negative triplet and the distance function of the $i$-th negative triplet. The weight of the negative sample $p(h'_i,r,t'_i)$ is defined as
\begin{equation}
  \begin{aligned}
    p((h'_i,r,t'_i)|(h,r,t)) = \frac{ \exp \alpha f_r(\mathbf{h'_i},\mathbf{t'_i})}{\sum_{j}\exp \alpha f_r(\mathbf{h'_j},\mathbf{t'_j})},
  \end{aligned}
\end{equation}
where $\alpha$ indicates the temperature of sampling.

\subsection{Theoretical Analysis}\label{sec4.2}
TransERR can model important relation patterns, including symmetry, antisymmetry, inversion, composition, subrelation and multiple. Please refer to Appendix~A for a detailed proof process.\\
\textbf{Proposition 1.} \emph{TransERR can infer the symmetry relation pattern.}\\
\textbf{Proposition 2.} \emph{TransERR can infer the antisymmetry relation pattern.}\\
\textbf{Proposition 3.} \emph{TransERR can infer the inversion relation pattern.}\\
\textbf{Proposition 4.} \emph{TransERR can infer the composition relation pattern.}\\
\textbf{Proposition 5.} \emph{TransERR can infer the subrelation pattern.}\\

\begin{table}[!htb]
	\centering
	\scalebox{0.8}{
		\begin{tabular}{cccccc}\hline
			\textbf{Dataset} & $|\mathcal{E}|$ & $|\mathcal{R}|$ & \textbf{\#Train} & \textbf{\#Valid} & \textbf{\#Test} \\ \hline
			ogbl-wikikg2     & 2,500k          & 535             & 16,109k          & 429k             & 598k            \\
			ogbl-biokg       & 94k             & 51              & 4763k            & 163k             & 163k            \\
			YAGO3-10         & 123k            & 37              & 1,079k           & 5k               & 5k              \\
			DB100K           & 100k            & 470             & 597k             & 50k              & 50k             \\
			FB15K            & 15k             & 237             & 483k             & 50k              & 50k             \\
			WN18             & 41k             & 18              & 141k             & 5k               & 5k              \\
			FB15K-237        & 15k             & 237             & 272k             & 18k              & 20k             \\
			WN18RR           & 41k             & 11              & 87k              & 3k               & 3k              \\
			Sports           & 1039            & 4               & 1312             & -                & 307             \\
			Location         & 445             & 5               & 384              & -                & 100             \\ \hline
		\end{tabular}
	}
	\caption{Statistics of the datasets in the experiment.}
	\label{tab:dataset}
\end{table}

\linespread{1.15}
\begin{table*}[!htb]
	\centering

	\scalebox{0.84}{\begin{tabular}{ccrcccrcc}
			\hline
			         & \multicolumn{4}{c}{\textbf{ogbl-wikikg2}} & \multicolumn{4}{c}{\textbf{ogbl-biokg}}                                                                                                        \\ \cmidrule(lr){2-5} \cmidrule(lr){6-9}
			         & \#Dim                                     & \#Params                                & Test MRR           & Valid MRR          & \#Dim & \#Params & Test MRR           & Valid MRR          \\ \hline
			TransE   & 500                                       & 1,250M                                & 0.4256             & 0.4272             & 2,000  & 187M  & 0.7452             & 0.7456             \\
			DistMult & 500                                       & 1,250M                                & 0.3729             & 0.3506             & 2,000  & 187M  & 0.8043             & 0.8055             \\
			ComplEx  & 250                                       & 1,250M                                & 0.4027             & 0.3759             & 1,000  & 187M  & 0.8095             & 0.8105             \\
			RotatE   & 250                                       & 1,250M                                & 0.4332             & 0.4353             & 1,000  & 187M  & 0.7989             & 0.7997             \\
			Rot\_Pro   & 200                                       & 1,000M                                & 0.5602             & 0.5740             & -  & -  & -             & -             \\
			PairRE   & 200                                       & 500M                                 & 0.5208             & 0.5423             & 2,000  & 187M  & {0.8164} & {0.8172} \\ 
			TripleRE   & 200                                       & 500M                                 & 0.5794             & 0.6045             & 2,000  & 187M  & {0.8191} & {0.8192} \\ 
			TranSHER   & 200                                       & 500M                                 & 0.5536             & 0.5662             & 2,000  & 187M  & \underline{0.8233} & \underline{0.8244} \\ \hline			
			TransERR  & 100                                       & \textbf{250M}                                 & \underline{0.6100} & \underline{0.6246} & 1,000  & \textbf{93M}   & 0.8153             & 0.8156             \\
			TransERR  & 200                                       & 500M                                 & \textbf{0.6359}    & \textbf{0.6518}    & 2,000  & 187M  & \textbf{0.8243}    & \textbf{0.8249}    \\ \hline
		\end{tabular}}
	\caption{Results on ogbl-wikikg2 and ogbl-biokg. Results are taken from the official leaderboard \cite{hu2020open}. The dashes mean that the results are not reported in the responding literature.}
	\label{tab:ogbl}
\end{table*}

\linespread{1.15}
\begin{table*}[!htb]
	\centering

	\scalebox{0.85}{
		\begin{tabular}{ccccccccccc}
			\hline
			                   & \multicolumn{5}{c}{\textbf{WN18RR}} & \multicolumn{5}{c}{\textbf{FB15K-237}}                                                                                                                                       \\ \cmidrule(lr){2-6} \cmidrule(lr){7-11}
			                   & MR                                  & MRR                                    & Hits@10        & Hits@3         & Hits@1         & MR           & MRR            & Hits@10        & Hits@3         & Hits@1         \\ \hline
			TransE $\heartsuit $   & 3384                                & 0.266                                  & 0.501          & -              & -              & 357          & 0.294          & 0.465          & -              & -              \\
			DistMult $\heartsuit $ & 5110                                & 0.43                                   & 0.49           & 0.44           & 0.39           & 254          & 0.241          & 0.419          & 0.263          & 0.155          \\
			ComplEx $\heartsuit $  & 5261                                & 0.44                                   & 0.51           & 0.46           & 0.41           & 339          & 0.247          & 0.428          & 0.275          & 0.158          \\
			TuckER             & -                                   & 0.470                                  & 0.526          & 0.482          & 0.443          & -            & 0.358          & 0.544          & 0.394          & \textbf{0.266} \\
			RotatE $\heartsuit $   & 3340                                & 0.476                                  & 0.571          & 0.492          & 0.428          & 177          & 0.338          & 0.533          & 0.375          & 0.241          \\
			Rotat3D    & 3328                                & 0.489                                  & 0.579          & 0.505          & 0.442          & 165          & 0.347          & 0.543          & 0.385          & 0.250          \\
			QuatE              & 3472                                & 0.481                                  & 0.564          & 0.500          & 0.436          & 176          & 0.311          & 0.495          & 0.342          & 0.221          \\
			Rot\_Pro           & 2815                                & 0.457                                  & 0.557          & 0.482          & 0.397          & 201          & 0.344          & 0.540          & 0.383          & 0.246          \\
			PairRE             &     -                               & -                                      & -              & -              & -              & 160          & 0.351          & 0.544          & 0.387          & 0.256          \\ 
			TripleRE   & -                                     & -                                   & -              & -              & -              & 142   & 0.351          & 0.544          & 0.387          & 0.256          \\
			TranSHER   & -                                     & -                                   & -              & -              & -              & -   & \textbf{0.360}          & 0.551          & \textbf{0.397}          & 0.264          \\  \hline
			TransERR            & \textbf{1167}                       & \textbf{0.501}                         & \textbf{0.605} & \textbf{0.520} & \textbf{0.450} & \textbf{125} & \textbf{0.360} & \textbf{0.555} & 0.396 & 0.264          \\ \hline
		\end{tabular}}
	\caption{Results on WN18RR and FB15K-237. Results of $\heartsuit $ are taken from  \citet{31sun2019rotate}. The best results are in bold. The dashes mean that the results are not reported in the responding literature.}
	\label{tab:wn18rr+fb237}
\end{table*}

\linespread{1.15}
\begin{table*}[!htb]
	\centering
	\scalebox{0.85}{
		\begin{tabular}{ccccccccccc}
			\hline
			                   & \multicolumn{5}{c}{\textbf{WN18}} & \multicolumn{5}{c}{\textbf{FB15K}}                                                                                                                                      \\ \cmidrule(lr){2-6} \cmidrule(lr){7-11}
			                   & MR                                & MRR                                & Hits@10        & Hits@3         & Hits@1         & MR          & MRR            & Hits@10        & Hits@3         & Hits@1         \\ \hline
			TransE $\heartsuit $   & -                                 & 0.495                              & 0.943          & 0.888          & 0.111          & -           & 0.463          & 0.749          & 0.578          & 0.297          \\
			DistMult $\heartsuit $ & 665                               & 0.797                              & 0.946          & -              & -              & 42          & 0.798          & 0.893          & -              & -              \\
			ComplEx $\heartsuit $  & -                                 & 0.941                              & 0.947          & 0.945          & 0.936          & -           & 0.692          & 0.840          & 0.759          & 0.599          \\
			TuckER             & -                                 & \textbf{0.953}                     & 0.958          & 0.955          & \textbf{0.949} & -           & 0.795          & 0.892          & 0.833          & 0.741          \\
			RotatE $\heartsuit $   & 309                               & 0.949                              & 0.959          & 0.952          & 0.944          & 40          & 0.797          & 0.884          & 0.830          & 0.746          \\
			Rotat3D    & 214                               & 0.951                              & 0.961          & 0.953          & 0.945          & 39          & 0.789          & 0.887          & 0.835          & 0.728          \\
			QuatE              & 338                               & 0.949                              & 0.960          & 0.954          & 0.941          & 41          & 0.770          & 0.878          & 0.821          & 0.700          \\
			PairRE             & 401                                 & 0.941                                  & 0.956              & 0.950              & 0.940              & 37 & 0.811          & \textbf{0.896} & 0.845          & 0.765          \\ 
			TripleRE   & -                                     & -                                   & -              & -              & -              & \textbf{35}   & 0.747          & 0.877          & 0.813          & 0.662          \\ \hline
			TransERR            & \textbf{82}                       & \textbf{0.953}                     & \textbf{0.965} & \textbf{0.957} & 0.945          & 41          & \textbf{0.815} & \textbf{0.896} & \textbf{0.848} & \textbf{0.767} \\ \hline
		\end{tabular}}
	\caption{Results on WN18 and FB15K. Results of $\heartsuit $ are taken from \citet{31sun2019rotate}.  Other results are taken from the corresponding original papers. The best results are in bold. The dashes mean that the results are not reported in the responding literature.}
	\label{tab:wn+fb}
\end{table*}

\section{Experimental Setup}

\subsection{Datasets}
We utilize the 10 most commonly utilized link prediction datasets and validate the effectiveness and generalizability of TransERR. We summarise the details of datasets in Table~\ref{tab:dataset}. ogbl-wikikg2~\cite{hu2020open} is a very large-scale dataset extracted from the Wikidata knowledge base~\cite{10.1145/2629489}. ogbl-biokg~\cite{hu2020open} consists massive biomedical data. YAGO3-10~\cite{6suchanek2007yago} is a subset of YAGO3, which are mainly from Wikipedia. DB100k~\cite{63ding2018improving} is a subset of DBpedia. FB15K~\cite{7bordes2013translating} is a subset of the Freebase knowledge base, while {FB15k-237} \cite{36toutanova2015observed} removes the inversion relations from FB15K. {WN18}~\cite{7bordes2013translating} is extracted from WordNet~\cite{5miller1995wordnet}. {WN18RR} \cite{11dettmers2018convolutional} removes inversion relations similar to FB15K-237. {Sports}~\cite{62wang2015knowledge} and {Location}~\cite{62wang2015knowledge} are both small-scale datasets. They are the main subrelation pattern. They are the NELL's~\cite{46mitchell2018never} subsets.

\subsection{Evaluation Protocol}

Following the SOTA methods \cite{31sun2019rotate,52chao2020pairre}, the quality of the ranking of each test triplet is evaluated via calculating all possible substitutions of head entity and tail entity : $(h',r,t)$ and $(h,r,t')$, where $h'$, $t'\in \mathcal{E}$. We evaluate the performance using the standard evaluation metrics, including Mean Rank (MR), Mean Reciprocal Rank (MRR) and Hits@N. Hits@N measures the percentage of correct entities in the top $n$ predictions. Higher values of MRR and Hits@N indicate better performance. Nevertheless, MR is the exact opposite of MRR and Hits@n. Hits@N ratio with cut-off values $N = 1, 3, 10$. For experiments on ogbl-wikikg2 and ogbl-biokg, we follow the evaluation protocol of these two benchmarks~\cite{hu2020open}.

\subsection{Implementation} 
We implement our proposed model via pytorch. We use Adam~\cite{adam} optimizer, and employ grid search to find the best hyperparameters based on the performance on the validation datasets. We report averaged test results across ten runs and use the random seeds from 0 to 9. We omit the variance as it is generally low.  We employ the official implementations~\cite{hu2020open} for ogbl-wikikg2\footnote{\url{https://ogb.stanford.edu/}}. In general, the embedding size $d$ is tuned amongst $\{100,200,500,1000,1500,2000\}$, the learning rate $\epsilon $ is tuned amongst $\{1e^{-3},5e^{-4},1e^{-4},5e^{-5},1e^{-5}\}$, the negative sample $n$ is selected in $\{128,256\}$, the self-adversarial sampling temperature $\alpha$ is selected from $\{0.5,1\}$, the fixed margin $\gamma$ are searched from 5 to 30.  All optimal hyperparameters are selected on the validation set.

\linespread{1.15}
\begin{table*}[!htb]
	\centering
	\scalebox{0.85}{
		\begin{tabular}{ccccccccccc}
			\hline
			         & \multicolumn{5}{c}{\textbf{YAGO3-10}} & \multicolumn{5}{c}{\textbf{DB100K}}                                                                                                                              \\ \cmidrule(lr){2-6} \cmidrule(lr){7-11}
			         & MR                                    & MRR                                 & Hits@10        & Hits@3         & Hits@1         & MR  & MRR            & Hits@10        & Hits@3         & Hits@1         \\ \hline
			TransE   & -                                     & -                                   & -              & -              & -              & -   & 0.111          & 0.270          & 0.164          & 0.016          \\
			DistMult & 5926                                  & 0.34                                & 0.54           & 0.38           & 0.24           & -   & 0.233          & 0.448          & 0.301          & 0.115          \\
			ComplEx  & 6351                                  & 0.36                                & 0.55           & 0.40           & 0.26           & -   & 0.242          & 0.440          & 0.312          & 0.126          \\
			ConvE    & 1671 & 0.44  & 0.62    & 0.49   & 0.35   & -   & -     & -       & -      & -      \\
			Rot\_Pro & 1797                                  & 0.542                               & 0.699          & 0.596          & 0.443          & 867   & 0.359              & 0.599              & 0.471              & 0.306              \\
			PairRE   & -                                     & -                                   & -              & -              & -              & -   & 0.412          & 0.600          & 0.472          & 0.309          \\ 
			TranSHER   & -                                     & -                                   & -              & -              & -              & -   & 0.431          & 0.589          & 0.476          & 0.345          \\ \hline
			TransERR  & \textbf{476}                          & \textbf{0.546}                      & \textbf{0.706} & \textbf{0.601} & \textbf{0.456} & \textbf{571} & \textbf{0.465} & \textbf{0.622} & \textbf{0.510} & \textbf{0.380} \\ \hline
		\end{tabular}}
	\caption{Results on YAGO3-10 and DB100K. Results are taken from the corresponding original papers. The best results are in bold. The dashes mean that the results are not reported in the responding literature.}
	\label{tab:yago+db}
\end{table*}

\linespread{1.15}
\begin{table*}[!htb]
	\centering

	\scalebox{0.85}{
		\begin{tabular}{ccccccccccccc}
			\hline
			                      & \multicolumn{5}{c}{\textbf{Sports}} & \multicolumn{5}{c}{\textbf{Location}}                                                                                                                             \\ \cmidrule(lr){2-6} \cmidrule(lr){7-11}
			                      & MR                                  & MRR                                   & Hits@10        & Hits@3         & Hits@1         & MR & MRR            & Hits@10        & Hits@3         & Hits@1         \\ \hline
			SimplE $^\spadesuit  $    & -                                   & 0.230                                 & 0.324          & 0.234          & 0.184          & -  & 0.190          & 0.315          & 0.210          & 0.130          \\
			SimplE$^{+\spadesuit}$  & -                                   & 0.404                                 & 0.508          & 0.440          & 0.394          & -  & 0.440          & 0.450          & 0.440          & 0.430          \\
			RotatE$^\diamondsuit $  & 191                                   & 0.420                                 & 0.535          & 0.503          & 0.420          & 73  & 0.486          & 0.550          & 0.480          & 0.455          \\
			PairRE $^\heartsuit $      & -                                   & 0.468                                 & -              & -              & 0.416          & -  & -              & -              & -              & -              \\ 
			TranSHER$^\diamondsuit $  & 151                                   & 0.479                                 & 0.537          & 0.509          & 0.433          & 71  & 0.475          & 0.575          & 0.470          & 0.435          \\ \hline
			TransERR               & \textbf{95}                                  & \textbf{0.499}                        & \textbf{0.570} & \textbf{0.526} & \textbf{0.447} & \textbf{30} & \textbf{0.563} & \textbf{0.645} & \textbf{0.565} & \textbf{0.520} \\\hline
		\end{tabular}}
	\caption{Results on Sports and Location. Results of $\spadesuit  $ and $\heartsuit $ are taken from \citet{58fatemi2019improved} and \citet{52chao2020pairre}, respectively. $\diamondsuit $ are obtained from our experiments. The best results are in bold. The dashes mean that the results are not reported in the responding literature.}
	\label{tab:sports+location}
\end{table*}

\section{Results and Analysis}\label{sec6}

\subsection{Main Results}\label{sec6.1}

To evaluate the effectiveness and the generalization of TransERR, we perform the experiments on 10 benchmark datasets of different scales. The results on ogbl-wikikg2 and ogbl-biokg are shown in Table~\ref{tab:ogbl}. TransERR achieves competitive results on large-scale datasets. On ogbl-wikikg2,  TransERR obtains significant improvements of $9.7\%$ than TripleRE with the same number of parameters (\#Dim 200) in Test MRR.  It is worth noting that TransERR still outperforms PairRE, TripleRE and TranSHER with fewer parameters (\#Dim 100). In addition, TransERR also outperforms RotatE and translation-based models with fewer parameters on ogbl-biokg. The results demonstrate that TransERR has a powerful capability to model large-scale KGs.

The comparison results for WN18RR and FB15K-237 are shown in Table~\ref{tab:wn18rr+fb237}. TransERR outperforms all the baselines in all metrics on WN18RR. Compared with RotatE, TransERR achieves improvements of $65.0\%$, $5.2\%$, $5.9\%$, $5.6\%$ and $5.1\%$, respectively. For FB15K-237, TransERR gains comparable results than existing distance-based models in MR, MRR and Hit@10. The above results confirm that TransERR has merit in modeling graph embeddings and improves the link prediction performance. This is because quaternions enable expressive rotation in the hypercomplex-valued space, and two unit quaternion vectors further narrow the translation distance between the head and tail entities.

The comparison results for WN18 and FB15K are shown in Table~\ref{tab:wn+fb}. We employ the same hyper-parameter settings and implementation compared with the previous works.  Since both the baselines and TransERR almost obtain the upper bound on Hits@N, the improvement of TransERR is still considered effective. We can see that TransERR achieves significant improvements on WN18 and FB15K.

The results on YAGO3-10 and DB100K datasets are shown in Table~\ref{tab:yago+db}. TransERR achieves the best results in all metrics for YAGO3-10. On DB100K, compared with the latest TranSHER, TransERR produces the optimal performance in all metrics except MR, which obtains significant improvements of $7.8\%$, $5.6\%$, $7.1\%$ and $10.1\%$, respectively.

We compare TransERR with the two latest methods SimplE$^+$ and PairRE, which are all focused on modeling subrelation pattern. As shown in Table \ref{tab:sports+location}, TransERR achieves the SOTA results on Sports and Location. Compared with the latest PairRE, TransERR obtains improvements of $6.6\%$ and $7.4\%$ in MRR and Hits@1 on Sports. For Location, TransERR outperforms all the baselines in all metrics.

	\begin{table*}[!htb]
		\centering
	
		\scalebox{0.8}{
			\begin{tabular}{ccccccccc}
				\hline \hline
				\textbf{Relation Type} & ~~1-to-1~~                                                 & ~~1-to-N~~                                                 & ~~N-to-1~~         & ~~N-to-N~~         & ~~1-to-1~~         & ~~1-to-N~~         & ~~N-to-1~ ~        & ~~N-to-N~~         \\ \hline
								   & \multicolumn{4}{c}{\textbf{Head prediction (MRR)}}     & \multicolumn{4}{c}{\textbf{Tail prediction (MRR)}}                                                                                                           \\ \cmidrule(lr){2-5} \cmidrule(lr){6-9}
				TransE             & 0.494                                                  & 0.456                                                  & 0.083          & 0.252          & 0.481          & 0.073          & 0.751          & 0.365          \\
				DistMult           & 0.213                                                  & 0.440                                                  & 0.071          & 0.229          & 0.212          & 0.053          & 0.727          & 0.345          \\
				ComplEx            & 0.356                                                  & 0.465                                                  & 0.091          & 0.247          & 0.373          & 0.062          & 0.741          & 0.356          \\
				RotatE             & 0.502                                                  & 0.465                                                  & 0.092          & 0.259          & 0.486          & 0.078          & 0.756          & 0.375          \\
				PairRE             & 0.493                                                  & 0.482                                                  & 0.116          & 0.271          & 0.492          & 0.071          & 0.775          & 0.386          \\
				TripleRE             & 0.499                                                  & 0.485                                                  & 0.116          & 0.283          & 0.493          & 0.077          & 0.776          & 0.388          \\
				TranSHER            & 0.501                                                  & 0.487                                                 & \textbf{0.119}          & \textbf{0.285}          & 0.494          & 0.079          & 0.779          & 0.389         \\
				TransERR            & \textbf{0.515}                                         & \textbf{0.495}                                         & 0.117 & 0.284 & \textbf{0.515} & \textbf{0.080} & \textbf{0.785} & \textbf{0.393} \\ \hline \hline
								   & \multicolumn{4}{c}{\textbf{Head prediction (Hits@10)}} & \multicolumn{4}{c}{\textbf{Tail prediction (Hits@10)}}                                                                                                       \\  \cmidrule(lr){2-5} \cmidrule(lr){6-9}
				TransE             & 0.494                                                  & 0.456                                                  & 0.083          & 0.252          & 0.481          & 0.073          & 0.751          & 0.365          \\
				DistMult           & 0.452                                                  & 0.640                                                  & 0.139          & 0.421          & 0.449          & 0.115          & 0.839          & 0.559          \\
				ComplEx            & 0.452                                                  & 0.643                                                  & 0.142          & 0.423          & 0.445          & 0.114          & 0.845          & 0.563          \\
				RotatE             & 0.601                                                  & 0.672                                                  & 0.175          & 0.468          & 0.586          & 0.143          & 0.880          & 0.615          \\
				PairRE             & 0.603                                                  & 0.670                                                  & 0.213          & 0.482          & 0.599          & 0.149          & 0.892          & 0.620          \\
				TripleRE             & 0.611                                                  & 0.671                                                  & 0.215          & 0.486          & 0.601          & 0.154          & 0.890          & 0.620          \\
				TranSHER             & 0.615                                                  & 0.674                                                  & 0.211          & 0.488          & 0.604          & \textbf{0.162}          & 0.891          & 0.624          \\
				TransERR            & \textbf{0.645}                                         & \textbf{0.696}                                         & \textbf{0.226} & \textbf{0.495} & \textbf{0.619} & 0.157 & \textbf{0.897} & \textbf{0.632} \\ \hline \hline
			\end{tabular}}
		\caption{Results on FB15k-237 by relation category.  Best results are in bold. Head prediction: predicting $h$ given $(?,r,t)$. Tail prediction: predicting $t$ given $(h,r,?)$.}
		\label{1-n}
	\end{table*}

	\begin{figure*}[t]
		\centering
		\subfloat[$\mathbf{r_1}$]{
			\label{a}
			\includegraphics[width=0.18\textwidth]{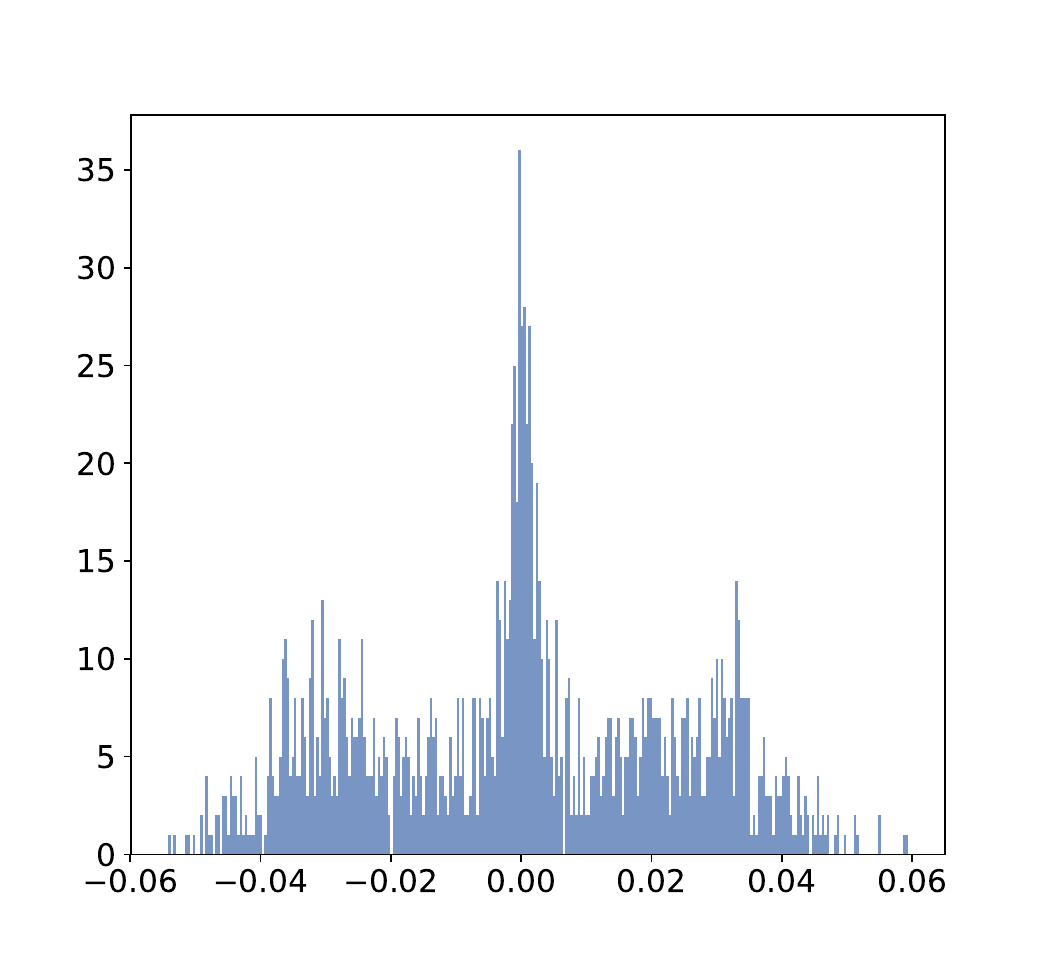}
		}\hfill
		\subfloat[$ \mathbf{r_1^{\vartriangleleft H}} +\mathbf{r_1^{\vartriangleleft T}}$]{
		\includegraphics[width=0.18\textwidth]{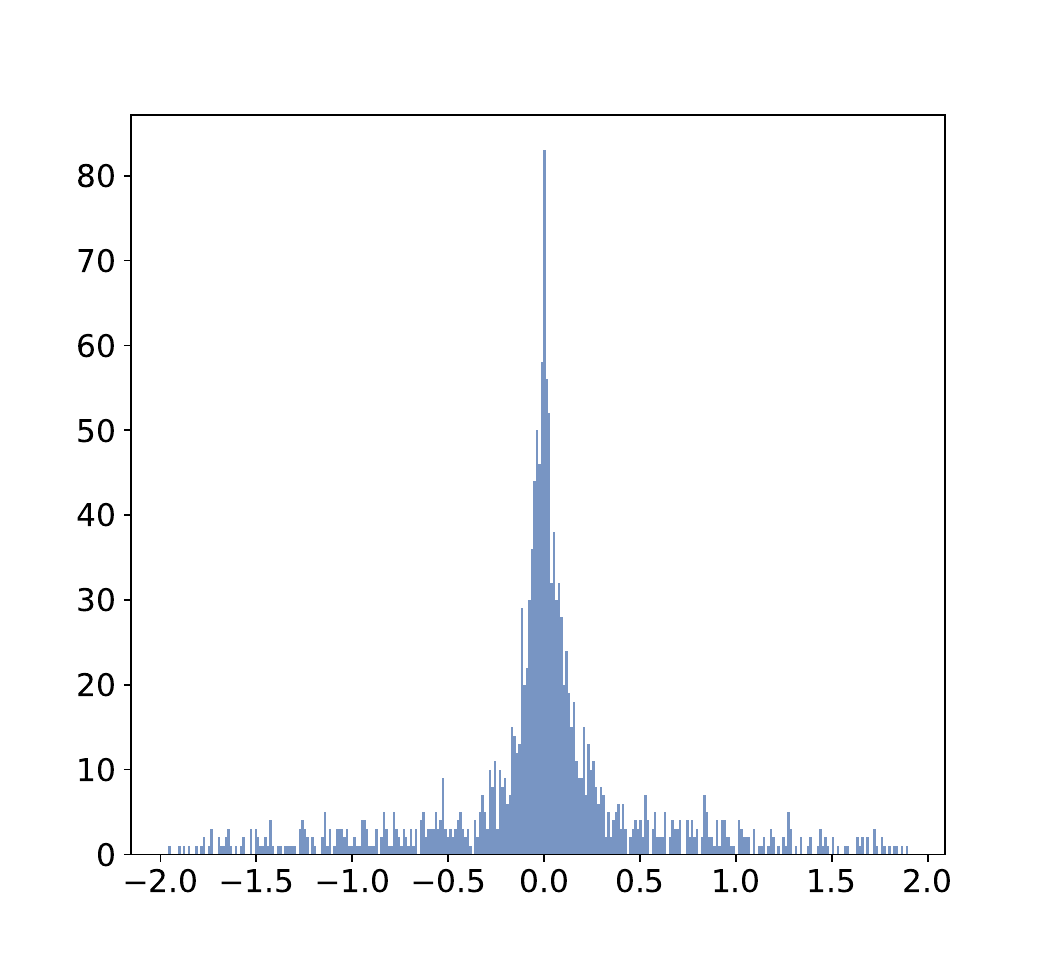}
		\label{b}
		}\hfill
		\subfloat[$\mathbf{r_2}$]{
			\includegraphics[width=0.18\textwidth]{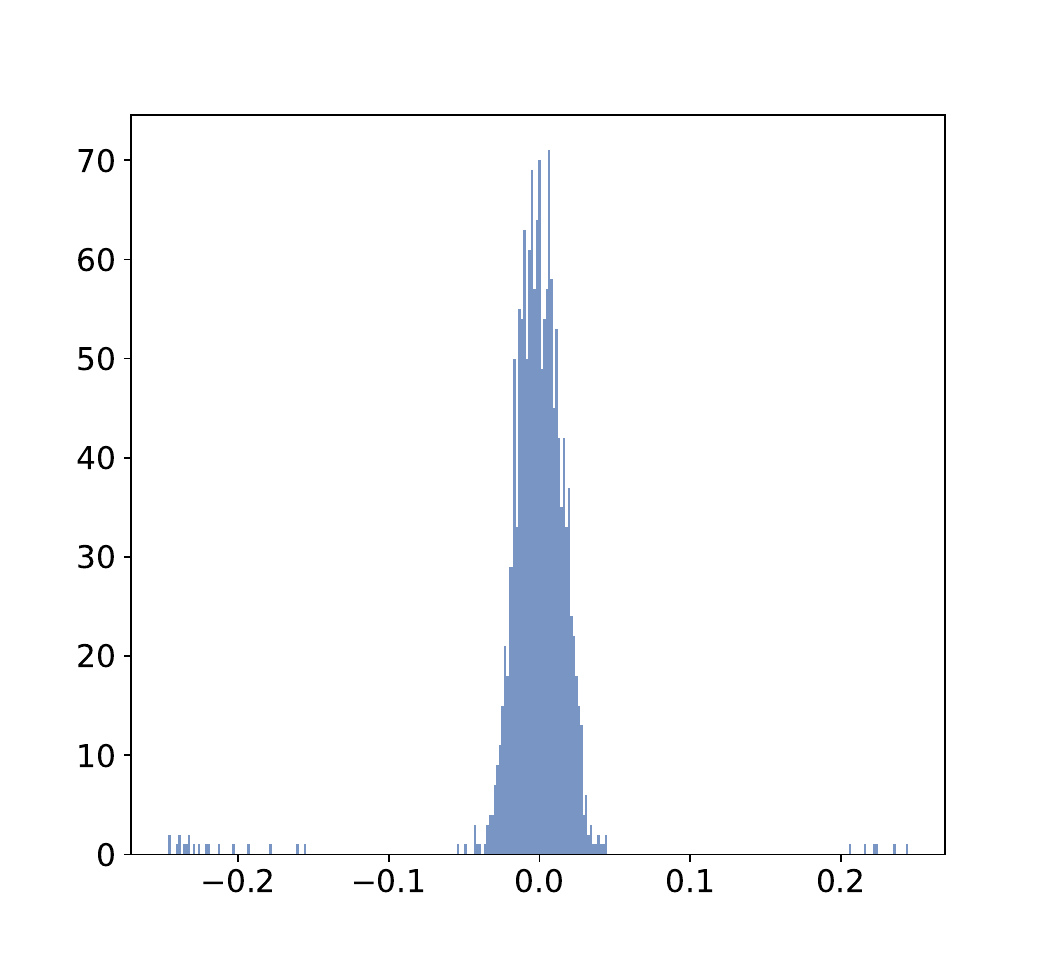}
			\label{c}
		}\hfill
		\subfloat[$ \mathbf{r_2^{\vartriangleleft H}} +\mathbf{r_2^{\vartriangleleft T}}$]{
		\includegraphics[width=0.18\textwidth]{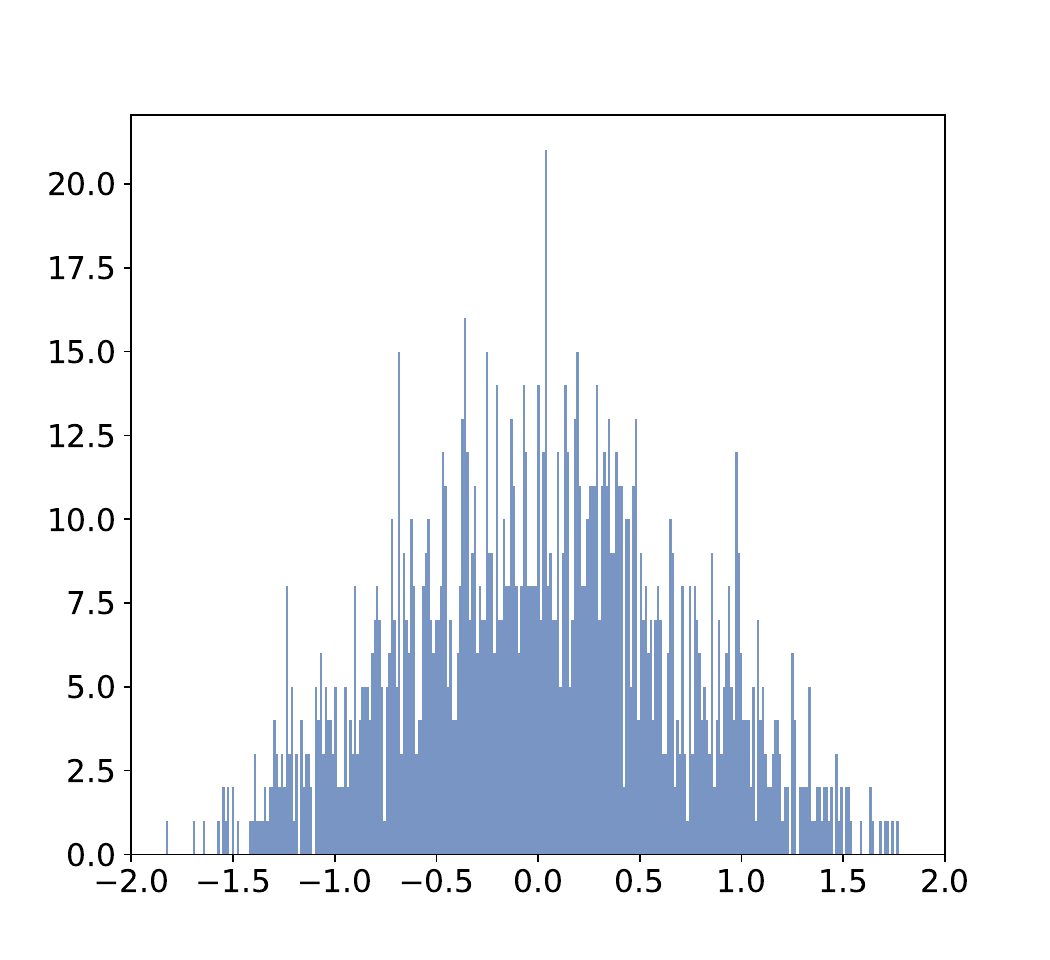}
		\label{d}
		}\hfill
		\subfloat[$\mathbf{r_3} $]{
			\includegraphics[width=0.18\textwidth]{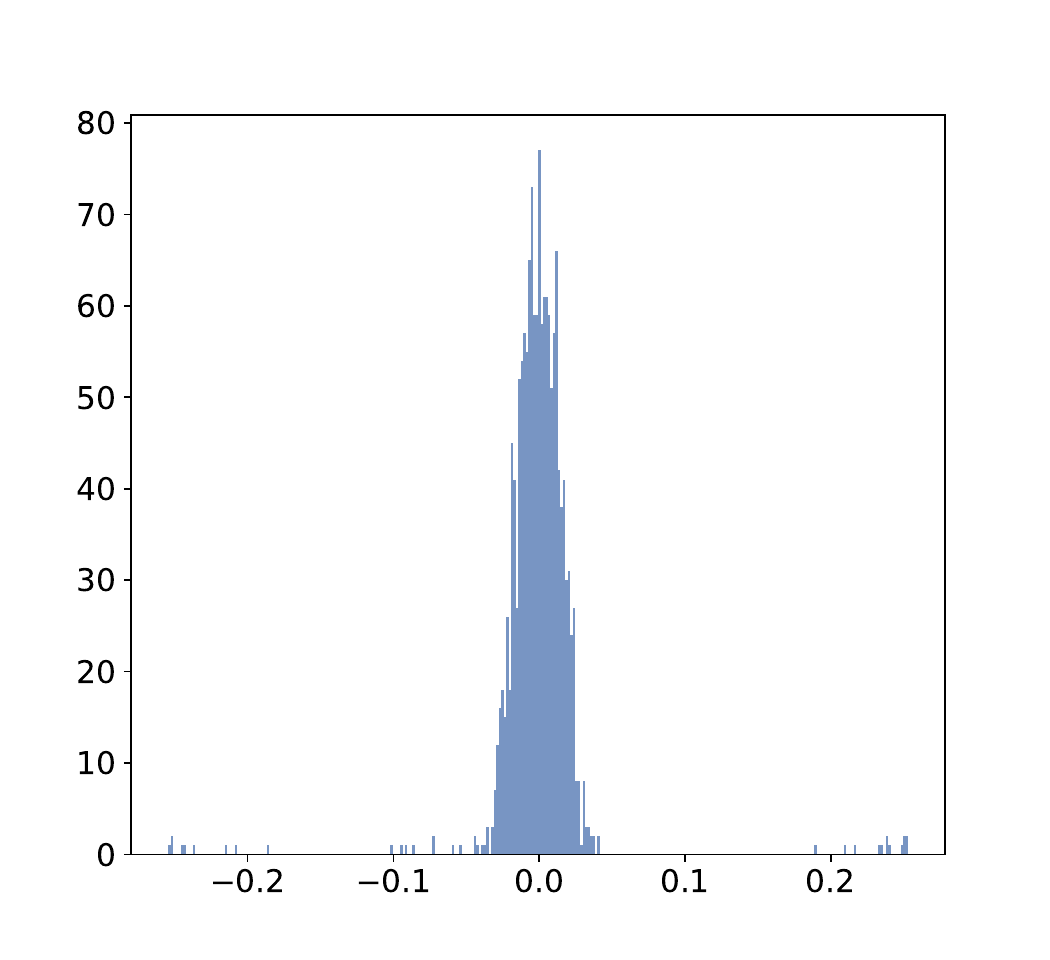}
			\label{e}
		}\hfill
		\subfloat[$\mathbf{r_2^{\vartriangleleft H}} \otimes \mathbf{r_3^{\vartriangleleft H}} - \mathbf{r_2^{\vartriangleleft T}} \otimes \mathbf{r_3^{\vartriangleleft T}} $]{
		\includegraphics[width=0.18\textwidth]{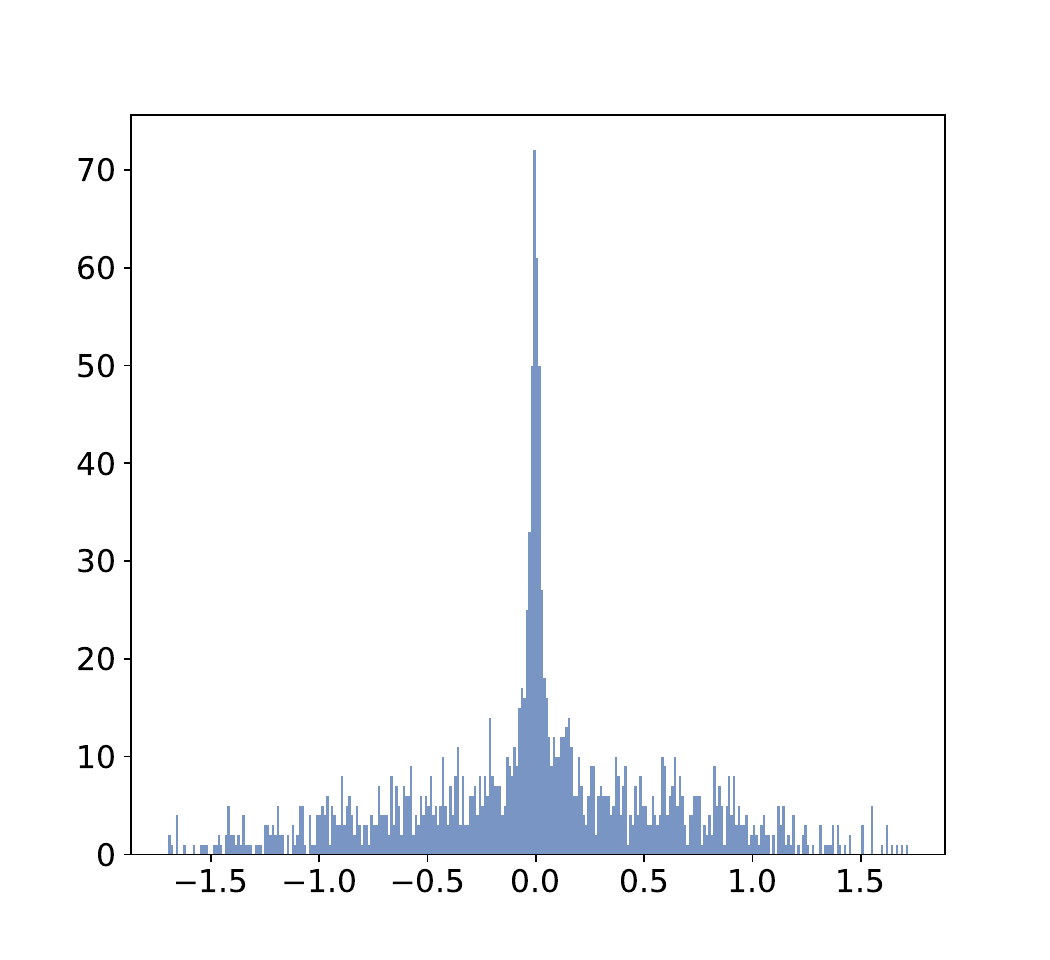}
		\label{f}
		}\hfill
		\subfloat[$\mathbf{r_2}+\mathbf{r_3} $]{
			\includegraphics[width=0.18\textwidth]{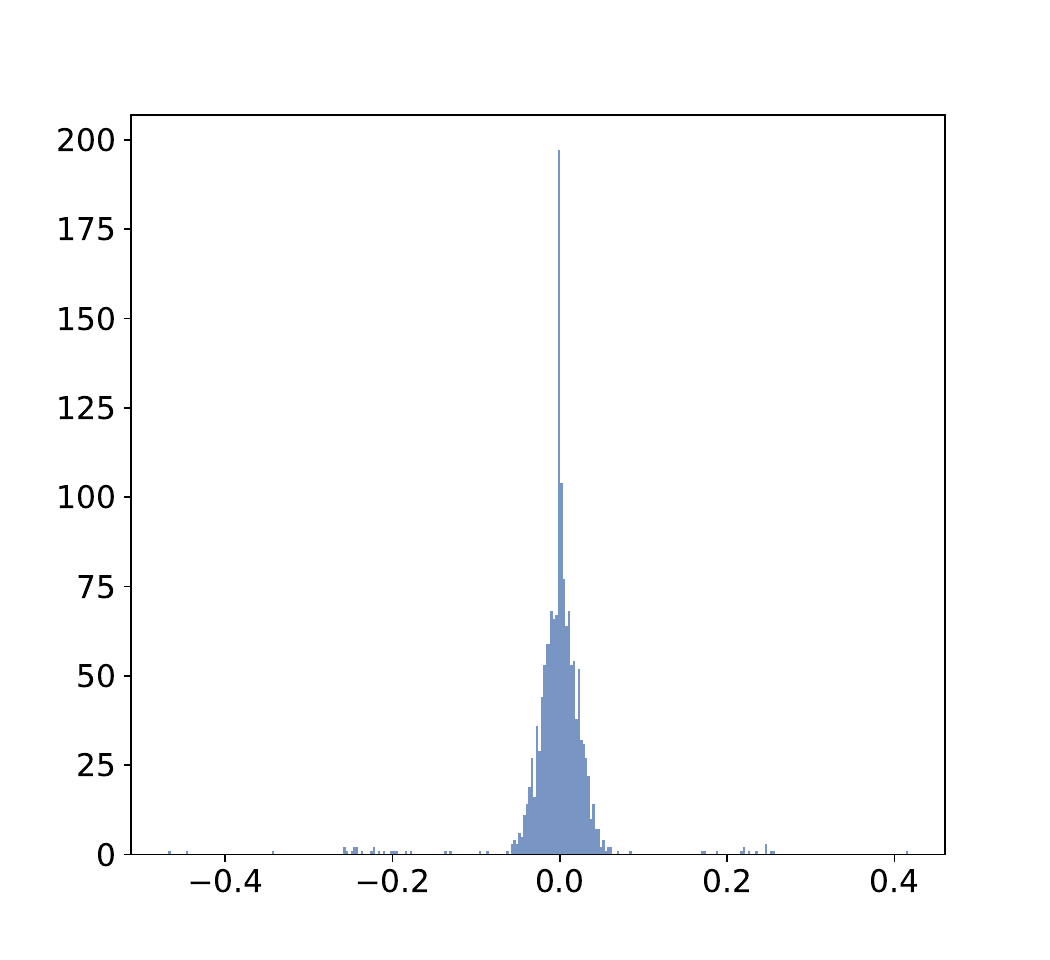}
			\label{g}
		}\hfill
		\subfloat[$\mathbf{r_4}$]{
			\includegraphics[width=0.18\textwidth]{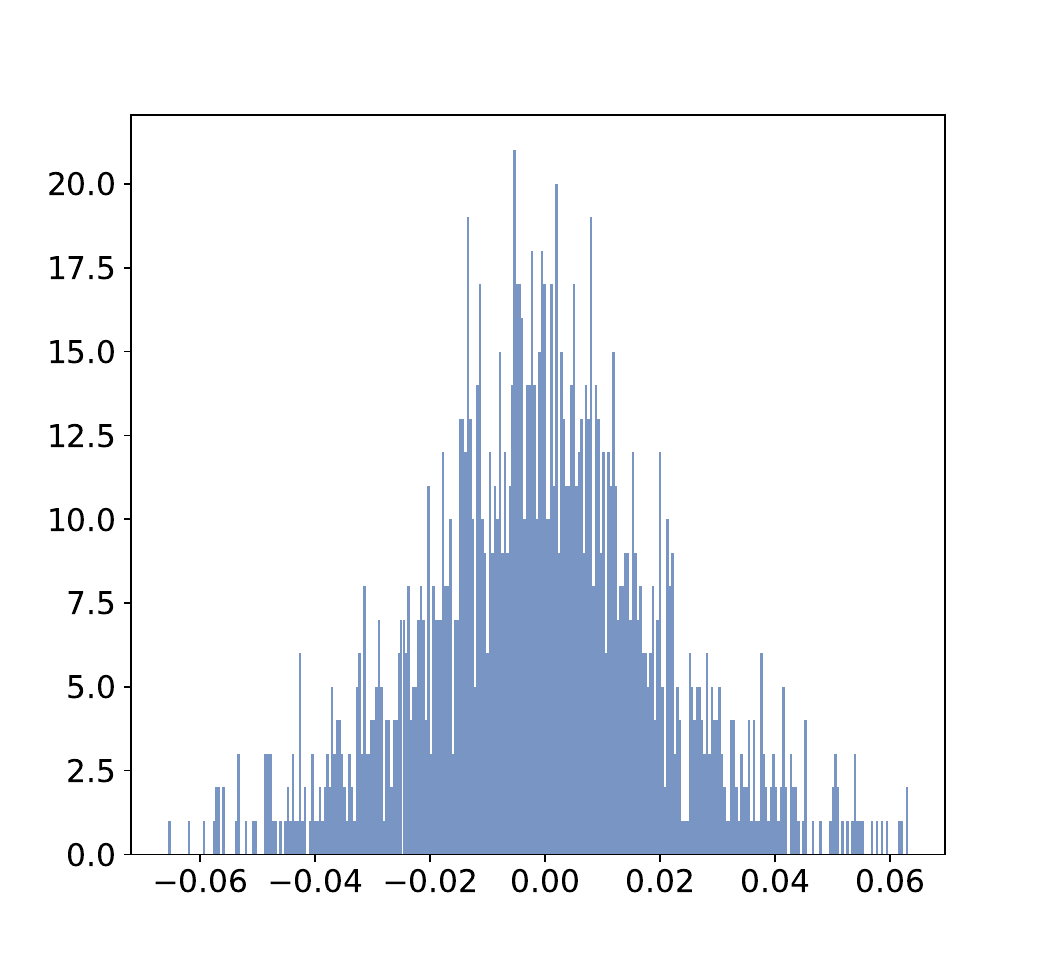}
			\label{h}
		}\hfill
		\subfloat[$\mathbf{r_5}$]{
			\includegraphics[width=0.18\textwidth]{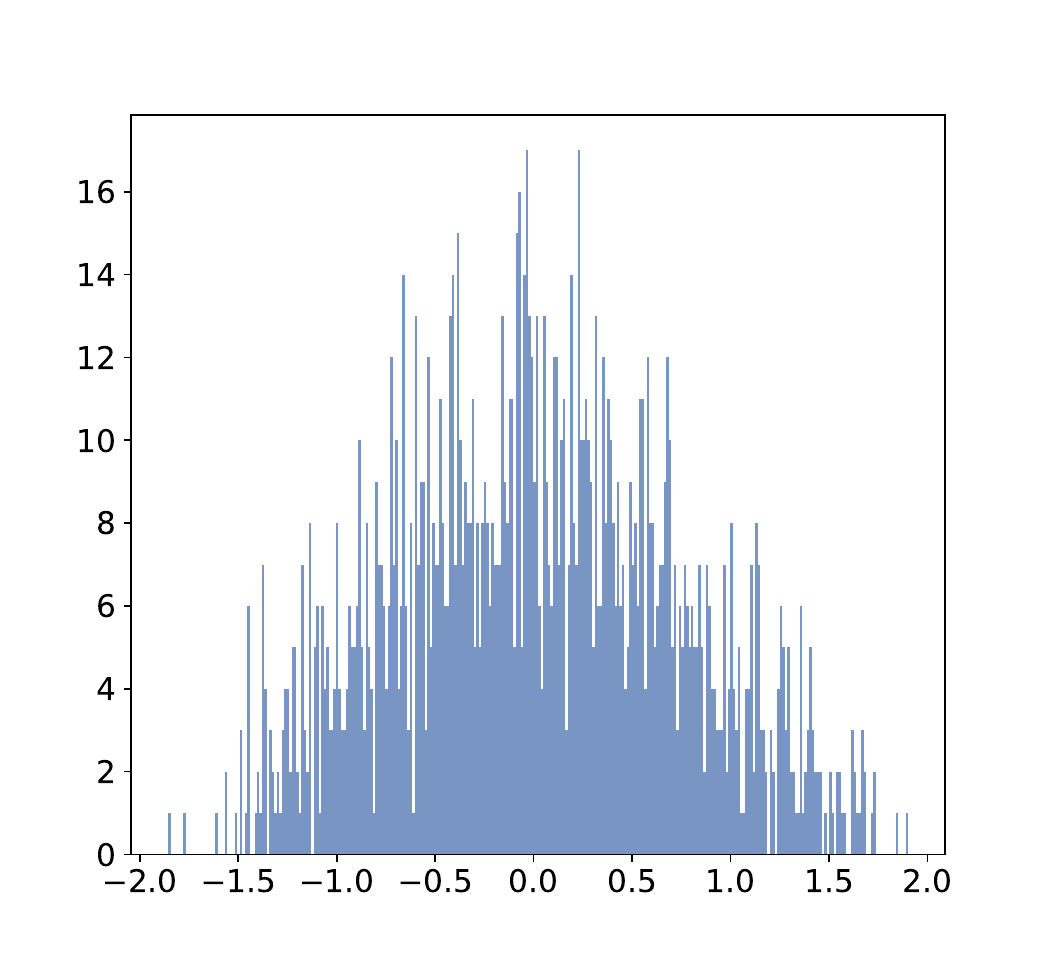}
			\label{i}
		}\hfill
		\subfloat[$\mathbf{r_6}$]{
			\includegraphics[width=0.18\textwidth]{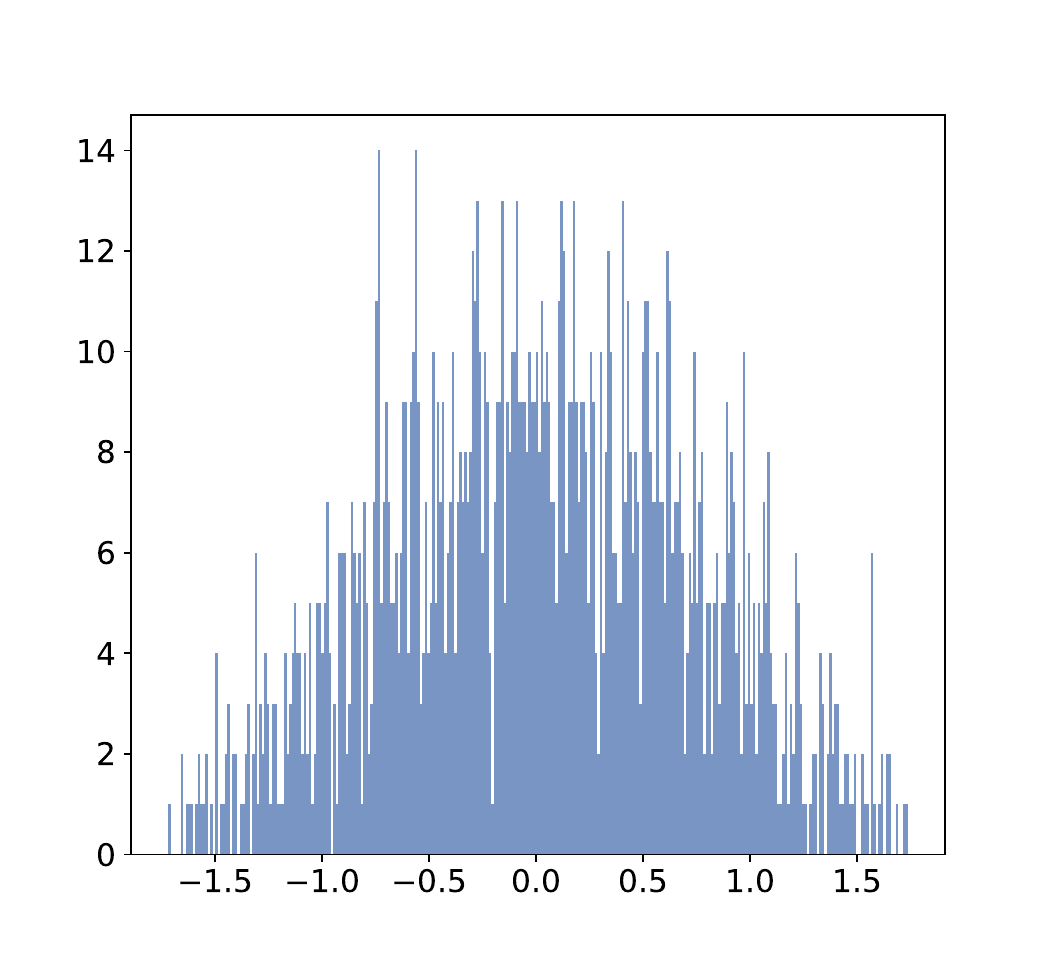}
			\label{j}
		}\hfill
		\subfloat[$\mathbf{r_6^{\vartriangleleft H}} -\mathbf{r_4^{\vartriangleleft H}} \otimes \mathbf{r_5^{\vartriangleleft H}}$]{
		\includegraphics[width=0.18\textwidth]{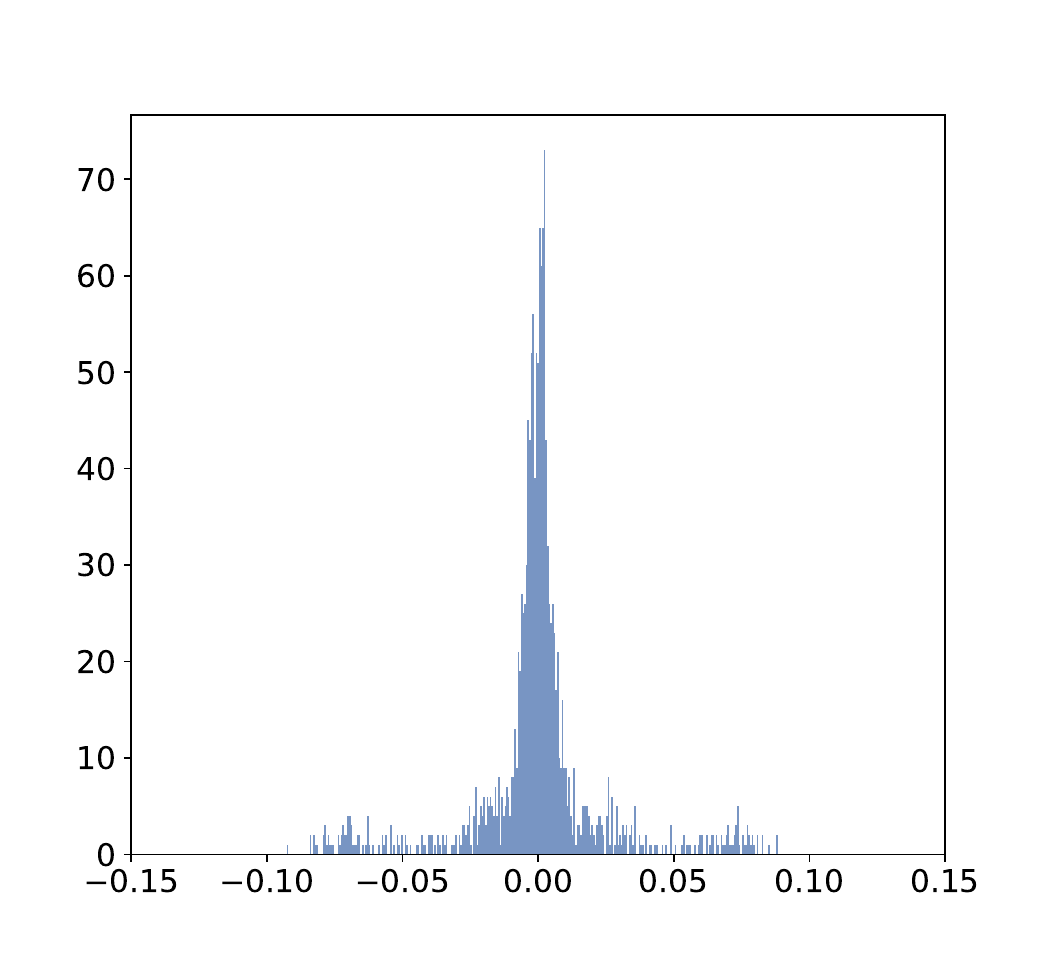}
		\label{k}
		}\hfill
		\subfloat[$\mathbf{r_6^{\vartriangleleft T}} -\mathbf{r_4^{\vartriangleleft T}} \otimes \mathbf{r_5^{\vartriangleleft T}}$]{
		\includegraphics[width=0.18\textwidth]{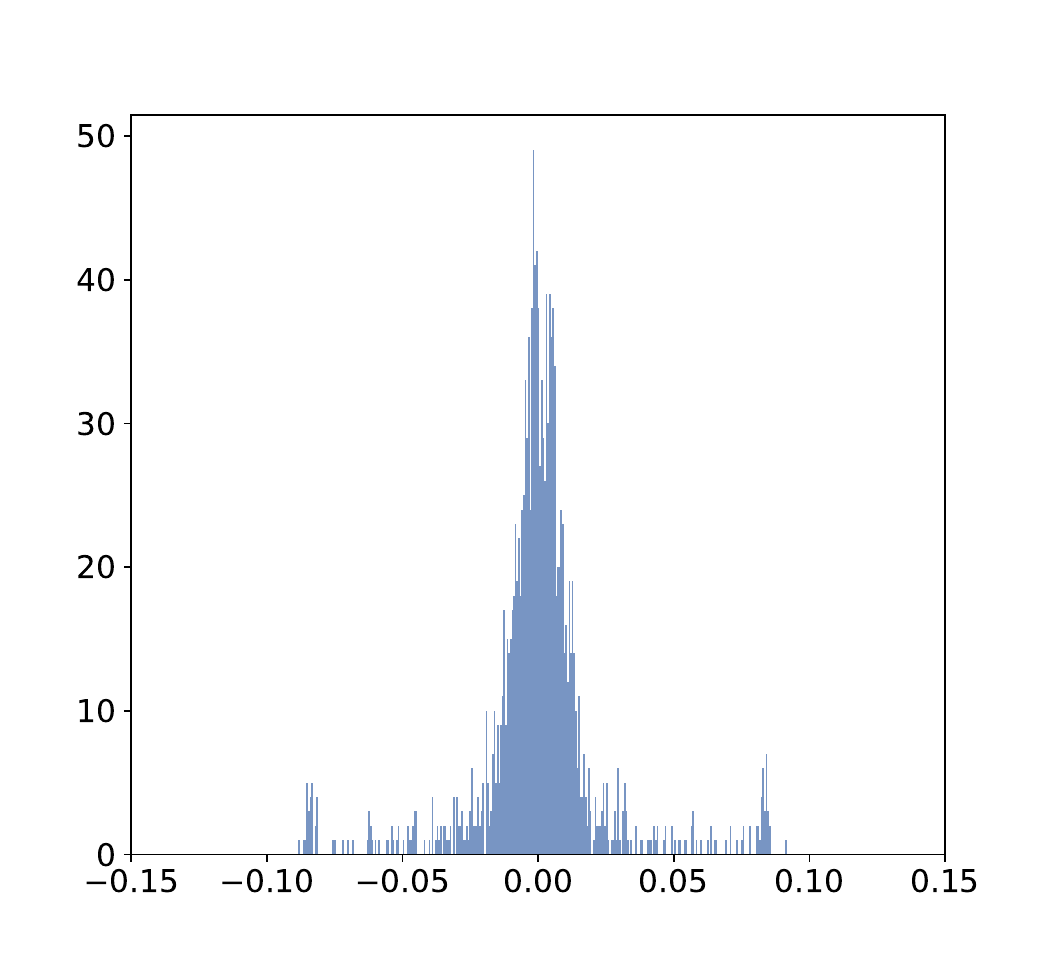}
		\label{l}
		}\hfill
		\subfloat[$\mathbf{r_6} - \mathbf{r_4} \otimes \mathbf{r_5^{\vartriangleleft H}}  - \mathbf{r_5}  \otimes\mathbf{r_4^{\vartriangleleft T}}$]{
		\includegraphics[width=0.18\textwidth]{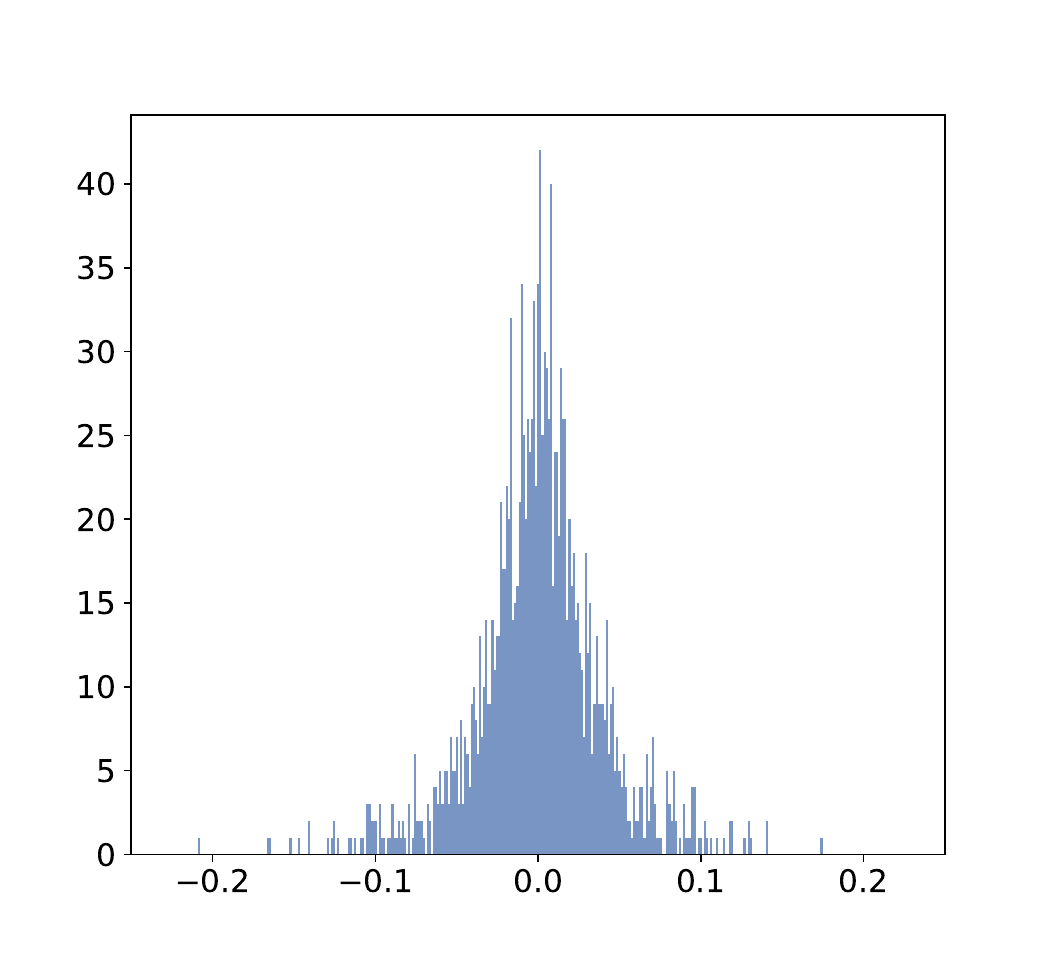}
		\label{m}
		}
		\caption{Relation embedding histograms for various relation patterns. $\mathbf{r_1}$ is \emph{/music/performance\_role/regular\_performances. /music/group\_membership/role}. $\mathbf{r_2}$ and $\mathbf{r_3}$  are \emph{/film/actor/film./film/performance/film} and  \emph{/film/film/starring./film/performance/actor}, respectively. $\mathbf{r_4}$, $\mathbf{r_5}$ and  $\mathbf{r_6}$ are \emph{/people/person/nationality}, \emph{/location/location/contains} and \emph{/people/person/place\_of\_birth}, respectively.}
		\label{fig:1-}
	\end{figure*}

	\subsection{Complex Relations Analysis}\label{sec6.3}
	This section analyzes the performance of the proposed TransERR on complex relations patterns. Following \citet{han2018openke}, we classify the relations into four categories: 1-to-1, 1-to-N, N-to-1 and N-to-N. The results of TransERR on different relation categories on FB15K-237 are shown in Table~\ref{1-n}. Compared with RotatE, TransERR obtains $2.5\%$ and  $7.3\%$ significant improvements in MRR and  Hits@10 for 1-1 scenario when predicting head entities. Besides, TransERR achieves $9.6\%$ and $5.7\%$ significant improvements in MRR and  Hits@10 for N-N scenario. The results prove that TransERR possesses a powerful ability to capture the latent information on complex relations (1-1, 1-N, N-1 and N-N) than the existing distance-based models.

	\subsection{Key Relation Patterns Analysis}\label{7-3}

	To illustrate the learned relation patterns that include symmetry, antisymmetry, inversion and composition, we visualize the histogram of relations embeddings in some examples in Figure~\ref{fig:1-}. The relations shown in Figure~\ref{fig:1-} are derived from FB15K.
	
	\textbf{Symmetry.} In Figure~\ref{a}, $\mathbf{r_1}$ is a symmetry relation. According to the theoretical analysis mentioned in Section~\ref{sec4.2}, TransERR can infer the symmetry relation pattern when $\mathbf{r_1^{\vartriangleleft H}} =-\mathbf{r_1^{\vartriangleleft T}}$ is satisfied. We can observe that $\mathbf{r_1^{\vartriangleleft H}}+\mathbf{r_1^{\vartriangleleft T}}$ tends to $0$ as much as possible in Figure~\ref{b}.
	
	\textbf{Antisymmetry.} On the contrary, $\mathbf{r_2}$ is an antisymmetric relation pattern in Figure~\ref{c}. TransERR encodes antisymmetry relation pattern when $\mathbf{r_2^{\vartriangleleft H}} \neq -\mathbf{r_2^{\vartriangleleft T}}$ is satisfied. We can find that Figure~\ref{d} satisfies the above condition.
	
	\textbf{Inversion.} $\mathbf{r_2}$ and $\mathbf{r_3}$ are an example of inversion relation pairs shown in Figure~\ref{c} and Figure~\ref{e}, respectively.  TransERR infers the inversion relation pattern when  $\mathbf{r_2^{\vartriangleleft H}} \otimes \mathbf{r_3^{\vartriangleleft H}} = \mathbf{r_2^{\vartriangleleft T}} \otimes \mathbf{r_3^{\vartriangleleft T}} $ and $\mathbf{r_2}= -\mathbf{r_3}$ are satisfied. Figure~\ref{f} and Figure~\ref{g} show that TransERR satisfies the above conditions.
	
	\textbf{Composition.}
	$\mathbf{r_4}$, $\mathbf{r_5}$, and $\mathbf{r_6}$ are an example of composition relation pattern shown in Figure~\ref{h}, Figure~\ref{i} and Figure~\ref{j}, respectively. TransERR infers the composition relation pattern when $\mathbf{r_6^{\vartriangleleft H}} =\mathbf{r_4^{\vartriangleleft H}} \otimes \mathbf{r_5^{\vartriangleleft H}}$, $\mathbf{r_6^{\vartriangleleft T}} =\mathbf{r_4^{\vartriangleleft T}} \otimes \mathbf{r_5^{\vartriangleleft T}}$ and $\mathbf{r_6} = \mathbf{r_4} \otimes \mathbf{r_5^{\vartriangleleft H}}  + \mathbf{r_5}  \otimes\mathbf{r_4^{\vartriangleleft T}}$  are satisfied. We can observe that Figure~\ref{k}, Figure~\ref{l} and Figure~\ref{m} satisfy the above conditions, respectively.
	
	\textbf{Subrelation.} To verify that TransERR can infer the subrelation pattern, we employ the hard rule constraint for subrelation. For $(e_1,r_1,e_2)$ and  $(e_1,r_2,e_2) $, we add the rules $(r1 \rightarrow  r2)$ are defined as follows:
	\begin{equation}
		\begin{aligned}
			\mathbf{r_2^{\vartriangleleft H}} = {\mathbf{r_1^{\vartriangleleft H}}} \mu, ~~\mathbf{r_2^{\vartriangleleft T}} = {\mathbf{r_1^{\vartriangleleft T}}} \mu, ~~\mathbf{r_2}={\mathbf{r_1}}  \mu.
		\end{aligned}
	\end{equation}

	We conduct experiments on Sports and add the following rules: \emph{CoachesTeam → PersonBelongsToOrganization} and  \emph{AthleteLedSportsTeam → AtheletePlaysForTeam}.
	The link prediction results are shown in Table~\ref{tab:sub-sports}. Compared with other models, which are focused on encoding subrelation pattern, TransERR achieves the SOTA results.  Compared with PairRE+Rule, TransERR+Rule obtains improvements of $6.9\%$ and $7.4\%$ in MRR and Hits@1. Results further show that TransERR can effectively model subrelation relation pattern.

	\linespread{1.15}
	\begin{table}[htb]
		\centering
		\scalebox{0.83}{
			\begin{tabular}{lcccc}
				\hline
							 & \multicolumn{4}{c}{\textbf{Sports}}                                                    \\ \cline{2-5}
							 & MRR                                 & Hits@10        & Hits@3         & Hits@1         \\ \hline
				SimplE$^+$   & 0.404                               & 0.508          & 0.440          & 0.394          \\
				PairE+Rule   & 0.475                               & -              & -              & 0.432          \\ 
				TransERR & 0.499                      & 0.570 & 0.526 & 0.447 			\\
				TransERR+Rule & \textbf{0.508}                      & \textbf{0.581} & \textbf{0.527} & \textbf{0.464} \\ \hline
			\end{tabular}}
		\caption{Link prediction results on Sports. All results are taken from the corresponding original papers. The dashes mean that the results are not reported in the responding literature.}
		\label{tab:sub-sports}
	\end{table}

	\begin{figure}[!htb]
		\centering
		\subfloat[CoachesTeam → PersonBelongsToOrganization]{
			\includegraphics[width=0.22\textwidth]{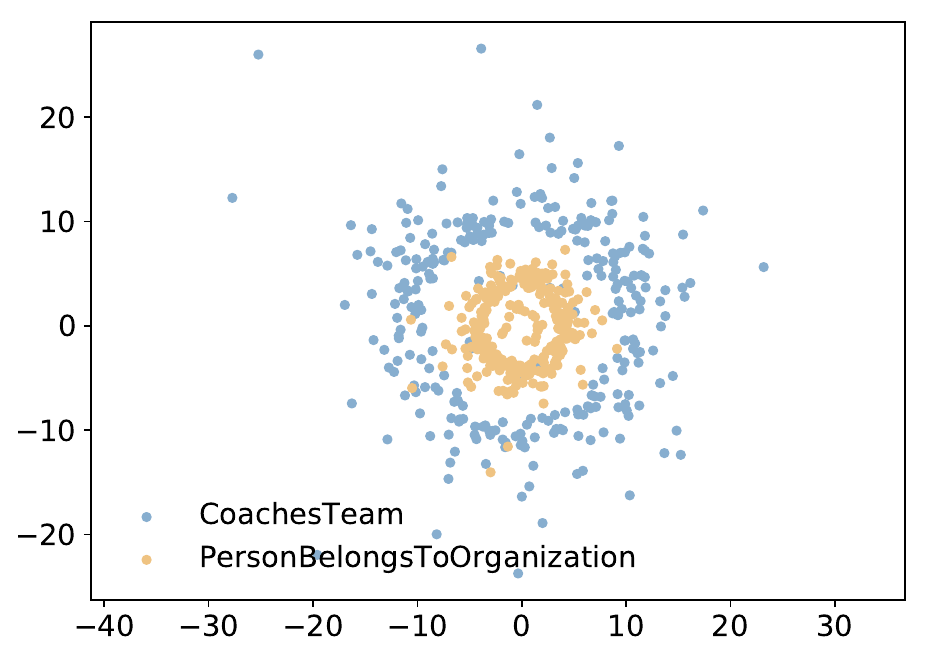}
		}\hfill
		\subfloat[AthleteLedSportsTeam → AtheletePlaysForTeam]{
			\includegraphics[width=0.22\textwidth]{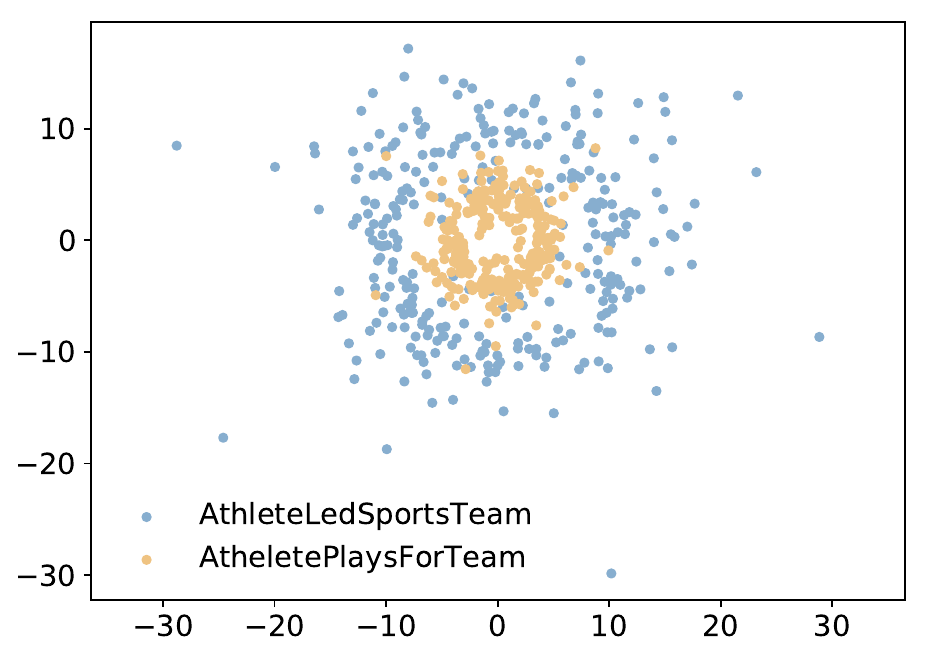}
		}\\
		\caption{Visualisation of relation embeddings  on Sports.}
		\label{fig:sub-relation}
	\end{figure}

	In addition, a quaternion number can be seen as a point on a 2D plane, and it contains a real part and three imaginary parts ($i$, $j$ and $k$).  We consider the three imaginary parts as a whole and plot the relation embeddings on a 2D plane.  Moreover, we have $\frac{\mathbf{r_2^{\vartriangleleft H}}}{\mathbf{r_1^{\vartriangleleft H}}} = \frac{\mathbf{r_2^{\vartriangleleft T}}}{\mathbf{r_1^{\vartriangleleft T}}} = \frac{\mathbf{r_2}}{\mathbf{r_1}} = \mu$. In this study, the optimal embedding dimension of Sports is $200$. Hence, the dimension of $r^{\vartriangleleft H}$, $r$ and $r^{\vartriangleleft T}$ are all $200$. Since the scale factor $\mu$ is the same between them, we plot $[r^{\vartriangleleft H}, r, r^{\vartriangleleft T}]$ and the embedding dimension is 600. That is, each relation contains 600 points. Furthermore, we employ the logarithmic scale to better display the differences between relation
	embeddings. The results are shown in Figure~\ref{fig:sub-relation}. We observe that TransERR can learn the hierarchy between relations and effectively model subrelation pattern with the hard rule constraint. In general, the above experimental analysis can once again prove the ability of our model in modeling key relation patterns.

	\linespread{1.15}
	\begin{table*}[t]
		\centering
		\scalebox{0.85}{
			\begin{tabular}{ccccccccc}
				\hline
																																							   & \multicolumn{2}{c}{\textbf{FB15K}} & \multicolumn{2}{c}{\textbf{FB15K-237}} & \multicolumn{2}{c}{\textbf{WN18}} & \multicolumn{2}{c}{\textbf{WN18RR}}                                                                      \\ \cmidrule(lr){2-3} \cmidrule(lr){4-5} \cmidrule(lr){6-7} \cmidrule(lr){8-9}
				\textbf{Scoring Function}                                                                                                                      & MRR                                & Hits@10                                & MRR                               & Hits@10                             & MRR            & Hits@10        & MRR            & Hits@10         \\ \hline
				$-\parallel \mathbf{h} \circ \mathbf{r^{ H}} + \mathbf{r}- \mathbf{t} \circ \mathbf{r^{  T}}   \parallel$  & 0.391                              & 0.561                                  & 0.235                             & 0.419                               & 0.909          & 0.944          & 0.469          & 0.563           \\
				$-\parallel \mathbf{h} \circ \mathbf{r^{\blacktriangleleft H}} + \mathbf{r}- \mathbf{t} \circ \mathbf{r^{\blacktriangleleft  T}}   \parallel$  & 0.732                              & 0.860                                  & 0.352                             & 0.537                               & 0.921          & 0.963          & 0.481          & 0.580           \\
				$-\parallel \mathbf{h}  \otimes \mathbf{r^{H}} + \mathbf{r}- \mathbf{t}  \otimes \mathbf{r^{T}}   \parallel$                                   & 0.425                              & 0.598                                  & 0.290                             & 0.454                               & 0.950          & 0.961          & 0.492          & 0.586           \\
				$-\parallel \mathbf{h}  \otimes \mathbf{r^{\vartriangleleft H}} + \mathbf{r}- \mathbf{t}  \otimes \mathbf{r^{\vartriangleleft T}}   \parallel$ & \textbf{0.815}                     & \textbf{0.896}                         & \textbf{0.360}                    & \textbf{0.555}                      & \textbf{0.953} & \textbf{0.965} & \textbf{0.501} & \textbf{ 0.605} \\ \hline
			\end{tabular}}
		\caption{Ablation of TransERR on FB15K, FB15K-237, WN18 and WN18RR. $\circ$ and $\otimes$ are Hadamard product and Hamilton product, respectively. $\mathbf{r^{ \blacktriangleleft H}}$ and $\mathbf{r^{\blacktriangleleft  T}}$ are normalized complex vectors. $\mathbf{r^{\vartriangleleft H}}$ and $\mathbf{r^{\vartriangleleft T}}$ are normalized quaternion vectors.}
		\label{abs}
	\end{table*}

\subsection{Ablation Study}\label{sec6.5}
In this part, in order to demonstrate that TransERR can better capture latent information between entity embeddings in the hypercomplex-valued space than in the complex-valued space, we encode TransERR in the complex space and conduct experiments on FB15K, FB15K-237, WN18 and WN18RR. As shown in the second and fourth rows of Table~\ref{abs},  we can observe that TransERR behaves better in the quaternion space. This is further evidence that TransERR can facilitate interaction information between the head and tail entities in the hypercomplex-valued space. Furthermore, we conduct an ablation study on quaternion normalization for TransERR, where rows one and two are a control group and rows three and four are a control group in Table~\ref{abs}. We remove the normalization step for $\mathbf{r^{H}}$ and $\mathbf{r^{T}}$ and utilize  $\mathbf{r^{H}}$ and $\mathbf{r^{T}}$ to rotate head entities and tail entities in the complex-valued space and in the hypercomplex-valued space, respectively. We conclude that the relational rotation's geometric property is lost, leading to poor performance. In addition, the unit quaternion has a higher degree of smooth rotation and spatial transformation ability than the unit complex number.

\section{Conclusion}
This paper proposes a simple yet effective distance-based KGE model (TransERR), which rotates entities with two normalized quaternion vectors in the hypercomplex-valued space. TransERR possesses a higher degree of translation freedom for graph embeddings. We provide formal mathematical proofs to demonstrate that TransERR can encode the key relation patterns.  Moreover, the results show that TransERR can effectively model complex relation patterns, including 1-1, 1-N, N-1 and N-N. The experiments also suggest that TransERR can maximize interaction information between entities in the hypercomplex-valued space. The experimental results fully illustrate the effectiveness and generalizability of our model. In addition, the results show that two unit quaternions can further narrow the distance between the head and tail entities, and thus avoid information loss in rotation.

\section{Acknowledgement}

This work was funded by National Natural Science Foundation of China (Grant No. 62366036), National Education Science Planning Project (Grant No. BIX230343), Key R\&D and Achievement Transformation Program of Inner Mongolia Autonomous Region (Grant No. 2022YFHH0077), The Central Government Fund for Promoting Local Scientific and Technological Development (Grant No. 2022ZY0198), Program for Young Talents of Science and Technology in Universities of Inner Mongolia Autonomous Region (Grant No. NJYT24033), Inner Mongolia Autonomous Region Science and Technology Planning Project (Grant No. 2023YFSH0017), Joint Fund of Scientific Research for the Public Hospitals of Inner Mongolia Academy of Medical Sciences (Grant No.2023GLLH0035 ), Natural Science Foundation of Inner Mongolia (Grant No. 2021BS06004).

\section{Bibliographical References}\label{sec:reference}
\bibliographystyle{lrec-coling2024-natbib}
\bibliography{lrec-coling2024-example}


\end{document}